\documentclass[letterpaper, 10 pt, conference]{ieeeconf}  

\IEEEoverridecommandlockouts                              

\usepackage{todonotes}
\usepackage{url}
\usepackage{xcolor}
\usepackage{pifont}
\usepackage{caption}
\usepackage{amsmath}

\usepackage{amsfonts}
\usepackage{amssymb}
\usepackage{algorithm}
\usepackage{algorithmic}
\usepackage{subcaption}

\usepackage{multirow}
\usepackage{makecell}

\newcommand{\cmark}{\ding{51}}%
\newcommand{\xmark}{\ding{55}}

\newcommand{\smark}{\ding{72}}

\newcommand{\vheading}[2][7em]{
	\rotatebox{90}{\parbox{#1}{\raggedright #2}}}

\title{\LARGE \bf
DeepRacer: Educational Autonomous Racing Platform for Experimentation with Sim2Real Reinforcement Learning
}

\author{Bharathan Balaji$^{1*}$,  Sunil Mallya$^{1*}$, Sahika Genc$^{1*}$, Saurabh Gupta$^{1}$, Leo Dirac$^{1}$, Vineet Khare$^{1}$, Gourav Roy$^{1}$\\
	 Tao Sun$^{1}$, Yunzhe Tao$^{1}$, Brian Townsend$^{1}$,  Eddie Calleja$^{1}$, Sunil Muralidhara$^{1}$, Dhanasekar Karuppasamy$^{1}$ 
\thanks{$^{1}$Authors are employees of Amazon Web Services. $*$ contributed equally. Send all correspondence to:
        {\tt\small bhabalaj@amazon.com}}%
}

\begin{document}

\maketitle
\thispagestyle{empty}
\pagestyle{empty}

\setcounter{footnote}{1}

\begin{abstract}

DeepRacer is a platform for end-to-end experimentation with RL and can be used to systematically investigate the key challenges in developing intelligent control systems.
Using the platform, we demonstrate how a 1/18th scale car can learn to drive autonomously using RL with a monocular camera. It is trained in simulation with no additional tuning in physical world and demonstrates: 1) formulation and solution of a robust reinforcement learning algorithm, 2) narrowing the reality gap through joint perception and dynamics, 3) distributed on-demand compute architecture for training optimal policies, and 4) a robust evaluation method to identify when to stop training. It is the first successful large-scale deployment of deep reinforcement learning on a robotic control agent that uses only raw camera images as observations and a model-free learning method to perform robust path planning. We open source our code and video demo on GitHub\footnote{DeepRacer training source code: \url{https://git.io/fjxoJ}}.
\end{abstract}

\section{Introduction}

Reinforcement Learning (RL) has been used to accomplish diverse robotic tasks: manipulation~\cite{andrychowicz2017hindsight,andrychowicz2018learning, gu2017deep,rusu17progressive}, locomotion~\cite{Hwangboeaau5872,xie2018feedback}, navigation~\cite{hsu2018distributed,choi2019deep,zhu2017target,kahn2018self}, flight~\cite{kim2004autonomous,sadeghi2016cad2rl}, interaction~\cite{chen2019crowd, christen2019guided}, motion planning~\cite{everett2018motion,sartoretti2019primal} and more. Due to high sample complexity and safety requirements, it is common to train the RL agent in simulation~\cite{andrychowicz2017hindsight,Hwangboeaau5872,brockman2016openai}. To reduce training time and encourage exploration, the agent is usually trained with distributed rollouts~\cite{fan2018surreal,pmlr-v87-liang18a,pmlr-v80-espeholt18a,pmlr-v80-liang18b}. For a successful transfer to the real world, researchers use calibration~\cite{andrychowicz2018learning,Tan-RSS-18}, domain randomization~\cite{peng2018sim, muratore2018domain, mandlekar2017adversarially, sadeghi2016cad2rl}, fine tuning with real world data~\cite{zhu2017target}, and learn features from a combination of simulation and real data~\cite{higgins2017darla,bharadhwaj2019data}.

To experiment with robotic reinforcement learning, one needs to have expertise in many areas, access to a physical robot, an accurate robot model for simulations, a distributed training mechanism and customizability of the training procedure such as modifying the neural network and the loss function or introducing noise. For the uninitiated, dealing with this complexity is daunting and dissuades adoption. As a result, much of prior work is limited to a single robot~\cite{andrychowicz2017hindsight,peng2018sim,Hightower:2017:KUR:3175917} or a few robots~\cite{sartoretti2019primal}. We reduce the learning curve and alleviate development effort with DeepRacer.

DeepRacer supports state-of-the-art deep RL algorithms~\cite{caspi_itai_2017_1134899}, simulations with the OpenAI Gym~\cite{brockman2016openai} interface, distributed rollouts and integration with cloud services. We introduce a training mechanism that decouples RL policy updates with the rollouts, which enables independent scaling of the simulation cluster and supports popular simulators such as Gazebo~\cite{Koenig-2004-394}. The DeepRacer 1/18th scale car is one realization of a physical robot in our platform that uses RL for navigating a race track with a fisheye lens camera. The car hardware includes GPU for executing the neural network policy locally, live streams the camera view over WiFi, the compute battery supports $\sim$6 hours of development time and retails at \$400. We have a corresponding robot model in simulation, along with rendering for multiple race tracks. We can train the RL policy with different simulation parameters and multiple tracks in parallel using our distributed rollout mechanism. 

We learn an end-to-end policy for navigating a race track. We use a single grayscale camera image as observation and discretized throttle/steering as actions. We train in simulation using the Proximal Policy Optimization (PPO) algorithm~\cite{schulman2017proximal}, which can converge in
\textless5 minutes and $\sim$5000 simulation steps. With no pre-processing, real world data or expert labeling, the learned policy successfully transfers from simulation to real tracks (sim2real~\cite{sadeghi2018sim2real}). The entire process from training a policy to testing in the real car takes \textless30 minutes. Multiple models can be trained in parallel with on-demand compute and stored in the car. Thousands of users have designed their own reward functions, trained their models on our platform, and demonstrated real track navigation. To the best of our knowledge, this is the first demonstration of model-free RL based sim2real at scale. 

DeepRacer serves as a testbed for many areas of RL research such as reducing sample complexity~\cite{kakade2003sample}, sim2real~\cite{jakobi1995noise} and generalizability~\cite{pmlr-v97-cobbe19a}. The car can log camera images, inertial sensor measurements, policy decisions. 
 Simulations can be randomized with different tracks, lighting, sensor and actuator noise. The learned policy can underfit/overfit to the simulation settings. We use a robust evaluation method to identify when the learned policy will generalize to the real world. We evaluate multiple checkpoints of the saved policy with domain randomization such as action noise and different starting points. Models that give good results in robust evaluation generalize well to the real world. Our policies trained with domain randomization generalize to multiple cars, tracks and to variations in speed, background, lighting, track shape, color and texture.  


\section{Related Work}


RL has been used in robotics for several decades~\cite{mataric1997reinforcement,asada1996purposive,gullapalli1994acquiring,mahadevan1992automatic}. Initial works used low dimensional state spaces due to scalability challenges. RL concepts were generalized to high dimensional problems with deep networks~\cite{mnih2015human,AlphaGo,schulman2015trust}. High variance, sample complexity and replicability challenges~\cite{henderson2018deep} in deep RL algorithms led to development of simulators~\cite{todorov2012mujoco}, benchmarks~\cite{brockman2016openai,tassa2018deepmind} and libraries~\cite{oguzsetting,dhariwal2017openai}. We build upon these works to create a platform for experimentation with simulation and real robots.

\textbf{Distributed Rollouts: } Algorithms that use distributed rollouts, where multiple simulations are executed in parallel to collect experience data, were introduced to reduce training time~\cite{andrychowicz2018learning,pmlr-v80-espeholt18a,mnih2016asynchronous}. 
OpenAI Baselines~\cite{dhariwal2017openai} uses OpenMPI~\cite{gabriel04:_open_mpi} to support distributed gradient algorithms, where each worker computes gradients on data collected. OpenAI Rapid~\cite{andrychowicz2018learning} generalizes it to a distributed system for the PPO algorithm and demonstrate sim2real transfer on dextrous manipulation. Flex~\cite{pmlr-v87-liang18a} extends the same distribution mechanism to use GPUs for simulation and hence can run 750 humanoid MuJoCo simulations with a single GPU. Chebotar et al.~\cite{chebotar2019closing} use Flex to demonstrate sim2real transfer for manipulation. Surreal~\cite{fan2018surreal} uses a decoupled rollout mechanism to support the experience replay algorithms, where each worker stores the experience data in a buffer and a separate training worker computes gradients. Ray RLlib~\cite{pmlr-v80-liang18b,moritz2018ray} introduces a stateful actor framework to support distributed rollouts. DeepRacer integrates with Intel Coach library~\cite{caspi_itai_2017_1134899} that supports \textgreater20 deep RL algorithms in an easy-to-use, modular interface. DeepRacer uses the same rollout mechanism as Surreal, and extends support for Gazebo. Similar to Rapid, DeepRacer can use different simulation settings for each worker and have separate evaluation workers that validate the performance of the current policy.

\textbf{Sim2Real: } Training RL policies in the real world is challenging due to high sample complexity and safety issues. Simulations alleviate these concerns and serve as a testbed to experiment with algorithms and debug software. However, sim2real transfer is challenging because of differences in dynamics, imagery and as simulated models are just approximations of the real world~\cite{peng2018sim, muratore2018domain, jakobi1995noise}. Domain randomization, where simulation parameters are perturbed during training, has been used for successful sim2real transfer for various robotic tasks~\cite{andrychowicz2018learning,sadeghi2016cad2rl,chebotar2019closing}. Methods include adding noise in dynamics~\cite{peng2018sim,andrychowicz2018learning} and imagery~\cite{sadeghi2016cad2rl,tobin2017domain}, learning model ensembles~\cite{mordatch2015ensemble,rajeswaran2016epopt}, adding adversarial noise~\cite{mandlekar2017adversarially,pinto2017robust} and assessing simulation bias~\cite{muratore2018domain}. Domain adaptation~\cite{patel2015visual} has also been used for sim2real, particularly to address the visual reality gap~\cite{higgins2017darla,bousmalis2018using,james2019sim,stein2018genesis}. 
DeepRacer serves as a platform to reproduce and experiment with sim2real methods. We demonstrate various forms of domain randomization in our experiments. Navigation with the DeepRacer car can be structured from simple, low speed, lane following to complex tasks such as high speed racing or commuting in traffic.  

Our distributed rollout mechanism facilitates iterative experimentation as policies converge faster and helps identify underfitting/overfitting. Prior sim2real works use a fixed number of simulation steps~\cite{andrychowicz2018learning,peng2018sim,tan2018sim,matas18adeformable}. We show that policies can both underfit and overfit to the simulation while training, as identified by prior works~\cite{muratore2018domain,pmlr-v97-cobbe19a,whiteson2011protecting}. We use a separate robust evaluation to identify the policy checkpoints that are likely to transfer well to the real world. 


\begin{table}[]
	\begin{center}
		\caption{Comparison of DeepRacer with contemporary RL and self-driving platforms. \smark indicates partial support.	\label{table:platform_comparison}}
		\begin{tabular}{|l|l|l|l|l|l|l|l|l|}
			\hline
			\textbf{Platform}   									  & \vheading{Simulation} & \vheading{Rigid Body Dynamics} & \vheading{Deep RL} & \vheading{Distributed Rollouts} & \vheading{Physical Robot} & \vheading{Sim2Real Demo} & \vheading{GPU on Robot} & \vheading{Robot Cost (USD)}  \\ \hline
			AutoRally~\cite{goldfain2019autorally}       & \cmark & \cmark & \cmark & \xmark & \cmark & \xmark & \cmark & 10K \\ \hline
			BARC~\cite{gonzales2016autonomous}     & \xmark & \xmark & \xmark & \xmark & \cmark & \xmark & \xmark & 500 \\ \hline
			Blue~\cite{gealy2019quasi}			              & \xmark & \xmark & \xmark & \xmark &  \cmark & \xmark & \xmark & 5K  \\ \hline
			CARLA~\cite{dosovitskiy17}                        & \cmark & \smark & \cmark & \cmark & \xmark & \cmark & \cmark & \xmark \\ \hline
			DonkeyCar~\cite{roscoe3donkey}              & \cmark & \smark & \cmark & \xmark & \cmark & \cmark & \xmark & 200 \\ \hline
			Duckietown~\cite{paull2017duckietown}    & \cmark & \smark & \cmark & \xmark & \cmark & \cmark & \xmark & 150 \\ \hline
			F1/10~\cite{o2019f1}                                    & \xmark & \xmark & \xmark & \xmark & \cmark & \xmark & \cmark & 3600 \\ \hline
			Fetch~\cite{andrychowicz2017hindsight,peng2018sim}	& \cmark & \cmark & \cmark & \cmark &  \cmark & \cmark & \xmark & 100K  \\ \hline
			Flex~\cite{pmlr-v87-liang18a}                     & \cmark & \cmark & \cmark & \cmark & \xmark & \cmark & \xmark & \xmark \\ \hline
			RACECAR~\cite{karaman2017project}       & \cmark & \cmark & \xmark & \xmark & \xmark & \xmark & \cmark & 2600 \\ \hline
			MuSHR~\cite{srinivasa2019mushr}            & \cmark & \xmark & \xmark & \xmark &  \cmark & \cmark & \cmark & 900 \\ \hline
			Poppy~\cite{lapeyre2014poppy}				 & \cmark & \cmark & \cmark & \cmark &  \cmark & \cmark & \xmark & 350 \\ \hline
			RLlib~\cite{pmlr-v80-liang18b}                   & \cmark & \cmark & \cmark & \cmark & \xmark & \xmark & \xmark & \xmark \\ \hline
			Surreal~\cite{fan2018surreal}                     & \cmark & \cmark & \cmark & \cmark & \xmark & \xmark & \xmark & \xmark \\ \hline
			\textbf{DeepRacer}							           & \cmark & \cmark & \cmark & \cmark & \cmark & \cmark & \cmark & 400 \\ \hline
		\end{tabular}
	\end{center}
	\vspace{-1em}
\end{table}

\textbf{Sim2Real Navigation:} Many works rely on simulators only for testing and use methods such as state estimation, motion planning and model predictive control (MPC)~\cite{paull2017duckietown,Boots-RSS-19,williams2017information} for navigation. Other works have used imitation learning, where expert demonstrations are given either by a person~\cite{roscoe3donkey,bojarski2016end} or with an MPC algorithm~\cite{pan2018agile}. Kahn et al.~\cite{kahn2018self} directly learn the RL policy in the real car, with a fixed maneuver when collision occurs. Domain randomization and image segmentation in simulations have been used to close the visual reality gap with a model based controller~\cite{sadeghi2016cad2rl,dosovitskiy17,loquercio2019deep}. Image pre-processing~\cite{zhang2019self}, learned embeddings~\cite{zhu2017target} and depth camera~\cite{wu2018learn} have been used to achieve sim2real transfer. Bharadhwaj et al.~\cite{bharadhwaj2019data} demonstrate sim2real transfer by mixing expert demonstrations with simulations. We observe that prior sim2real works rely on a model based controller for high speed navigation~\cite{loquercio2019deep,drews2019vision} or achieve slow speeds 
because of poor transfer of dynamics~\cite{zhang2019self,wu2018learn}. With DeepRacer, we demonstrate speeds of 1.6m/s with a single grayscale monocular image as input and discretized steering/throttle as output. We use simple, non-recurrent networks for our policy and still demonstrate robustness in the real world to multiple cars, tracks, and variations in the environment. We also achieve slow speed (0.5m/s) sim2real transfer with \textless5 minutes of training.

Table \ref{table:platform_comparison} compares DeepRacer with other platforms for RL, sim2real and autonomous driving. The other simulation platforms can also be used with DeepRacer. 
We provide an easy-to-use, economical and flexible platform with support for distributed RL, domain randomization and robust evaluation. DeepRacer tools have enabled us to replicate sim2real RL policy transfer with consistency and at scale.

\section{Autonomous Racing with RL}

\begin{figure}
	\centering
	\includegraphics[width=\linewidth]{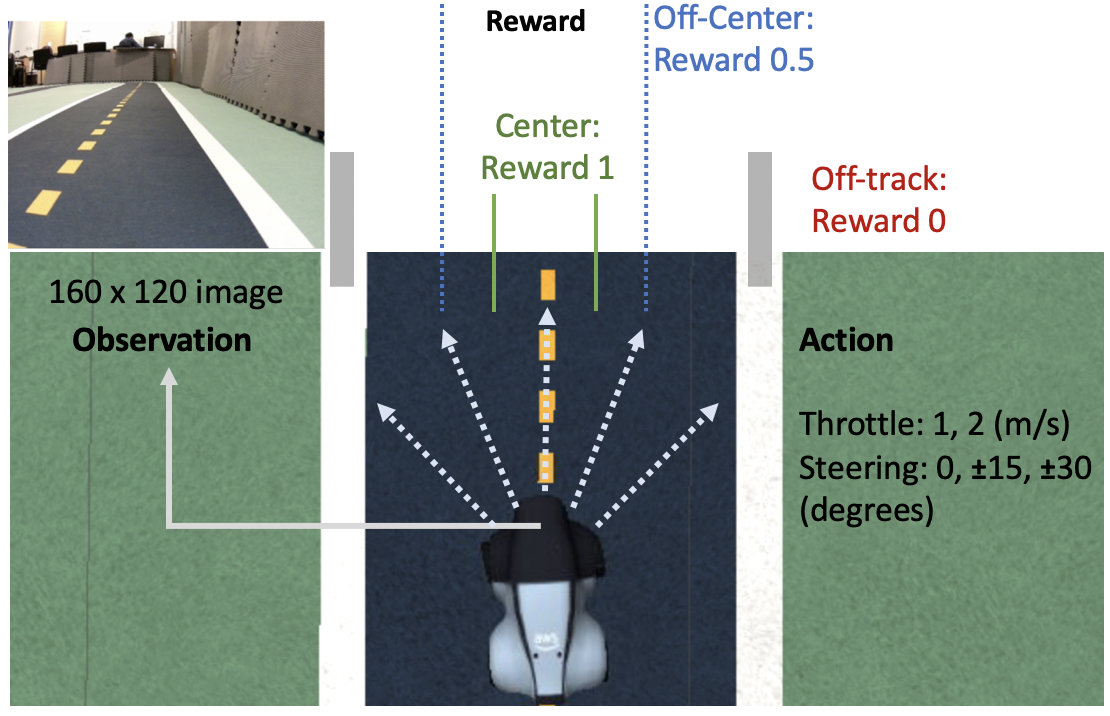}
	\caption{Observation, action and reward for DeepRacer agent}
	\vspace{-1em}
	\label{fig:state_action_reward}
\end{figure}

In our formulation, the agent steers the car and the environment is the race track. The track is marked by white lanes, there is a single car on track with no obstacles and the car only moves forwards. The image from the car's camera is the observation, and actions are the throttle/steering of the car. As the agent does not receive the full state such as the the track layout, this is a partially observed Markov Decision Process. An episode starts with the car somewhere on track and finishes when the car goes off-track or finishes a lap.

The images from the camera are streamed at 15 fps, downsized to 160 x 120 pixels and converted to grayscale. We discretize the actions to 10 values, with 2 levels for throttle and 5 for steering. Users can customize this discretization, which get mapped to low level controls. We fix the maximum throttle in simulation and set it manually in the real car. We incentivize the agent to stay close to the center line of the track. If the car is at the edge of the track, a small deviation can off-road the car and the track is not visible in the image. Staying close to the center of the track leads to a stable policy. Users can customize this reward function. Figure \ref{fig:state_action_reward} illustrates our problem formulation.


\subsection{Reinforcement Learning Algorithm}

%
We use PPO, a state-of-the-art policy gradient algorithm~\cite{schulman2017proximal}. The algorithm uses two neural networks during training -- a policy network and a value network. The policy network decides which action to take given an image as input and the value network estimates the expected cumulative discounted reward given the image. The agent initializes a policy that takes random actions. The policy network is used to interact with the simulation environment to collect data. The resulting dataset is used to update the policy and value networks as per the algorithm's loss function. The updated policy is used to interact with the environment to collect more data and the training cycle continues until a time limit.

The policy loss function maximizes the actions that give higher rewards on average as given by the generalized advantage estimation algorithm~\cite{schulmanetal_ICLR2016} and applies a clipped importance sampling weight as the policy that collects the dataset is an older version of the policy being updated. The value loss function uses the mean squared error between the predicted value and the observed value. Only the policy network gets deployed in the real car. By default, we use three convolutional layers and two fully connected layers for both networks. We train a new policy every 20 episodes. The full list of hyperparameters is given in our source code.

\section{DeepRacer Design and Implementation}


We decouple the simulation data collection from the policy updates. We use RoboMaker~\cite{robomaker} for our simulations with Gazebo and SageMaker~\cite{sagemaker} to train our policy with the RL Coach~\cite{caspi_itai_2017_1134899} library. Simulations help us train without manual effort. The decoupled training allows us to use separate machines which are specialized for simulations (e.g. license, Mac/Windows OS) and neural network training (e.g. GPU, large RAM) respectively. We also get the flexibility to launch multiple simulations each with their own settings for domain randomization as well as evaluate policies in parallel.

\subsection{Training Workflow}

\begin{figure}[t]
	\centering
	\includegraphics[width=\linewidth]{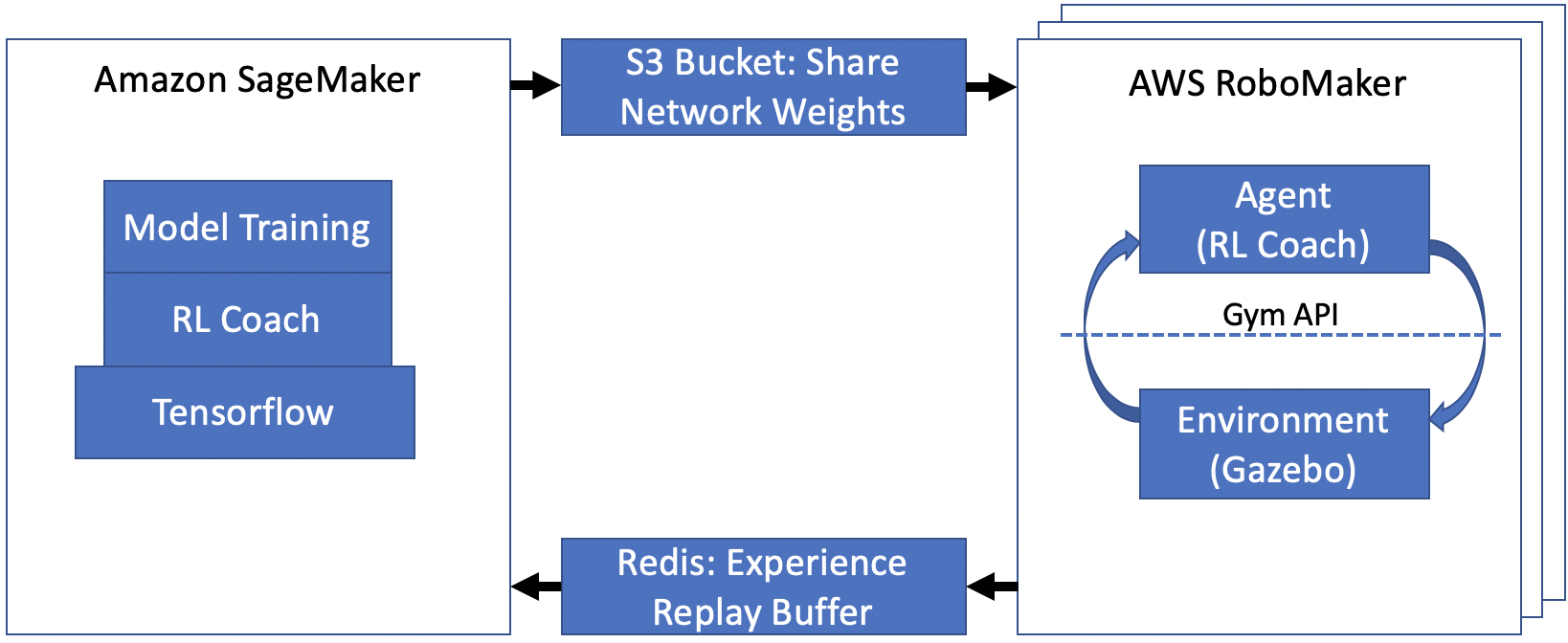}
	\caption{Training the agent with DeepRacer distributed rollouts}
	\label{training_ppo}
	\vspace{-1em}
\end{figure}

Figure \ref{training_ppo} shows the DeepRacer training workflow. The training starts by initializing the policy/value network models and hyper-parameters in SageMaker. The neural network models are saved in S3~\cite{s3}, an object store service. RoboMaker initializes the simulation, the agent and loads the models from S3. The agent interacts with the simulation over the OpenAI Gym interface. The agent takes actions $a$ (steering/throttle) based on the observation $o$ (camera image). The simulator updates the position of the car based on the action and returns with the updated camera image and reward $r$. The experiences collected in the form of $\langle o_t, a_t, r_t, o_{t+1} \rangle$ are stored in Redis~\cite{redis}, an in-memory database. SageMaker trains the neural networks with data collected in Redis and saves the models in S3. RoboMaker copies the model from S3 and creates more experience data. The cycle continues until training stops. The models in S3 are continually evaluated in a separate simulation to assess convergence and generalizability. Models in S3 can be deployed on the real car. While we show our results with the PPO algorithm, our architecture can be used for various experience replay based algorithms such as DQN~\cite{mnih2015human}, DDPG~\cite{lillicrapHPHETS15} and SAC~\cite{haarnoja18sac}. Robomaker can be replaced with other simulators that can integrate with the Gym interface.

\begin{figure}
	\centering
	\begin{subfigure}[b]{\columnwidth}
		\centering
		\includegraphics[width=\columnwidth]{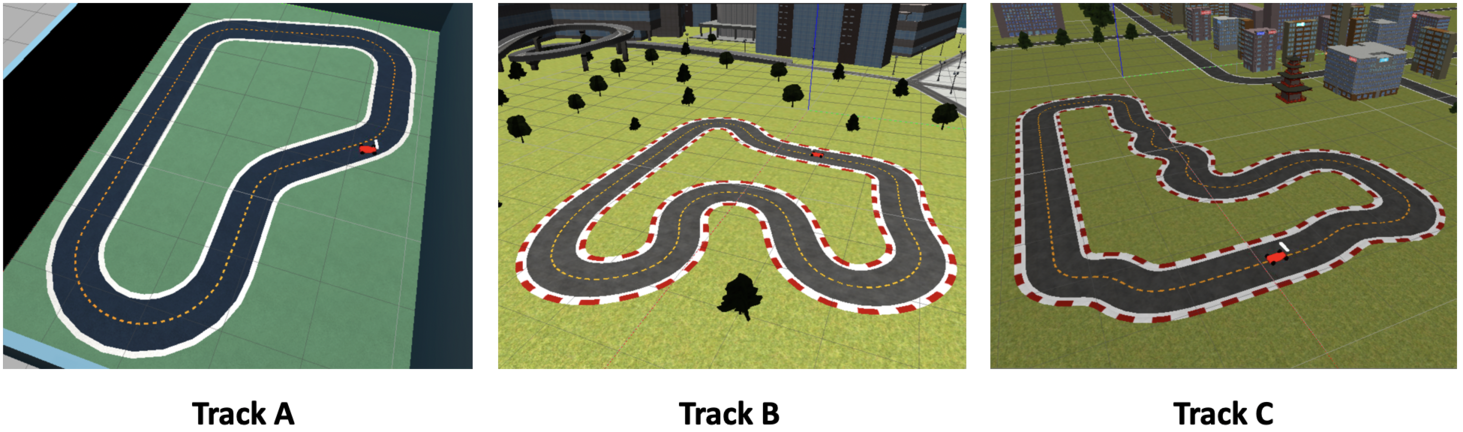}
		\caption{Simulation tracks}
	\end{subfigure}
	\hfill
	\begin{subfigure}[b]{\columnwidth}
		\centering
		\includegraphics[width=\columnwidth]{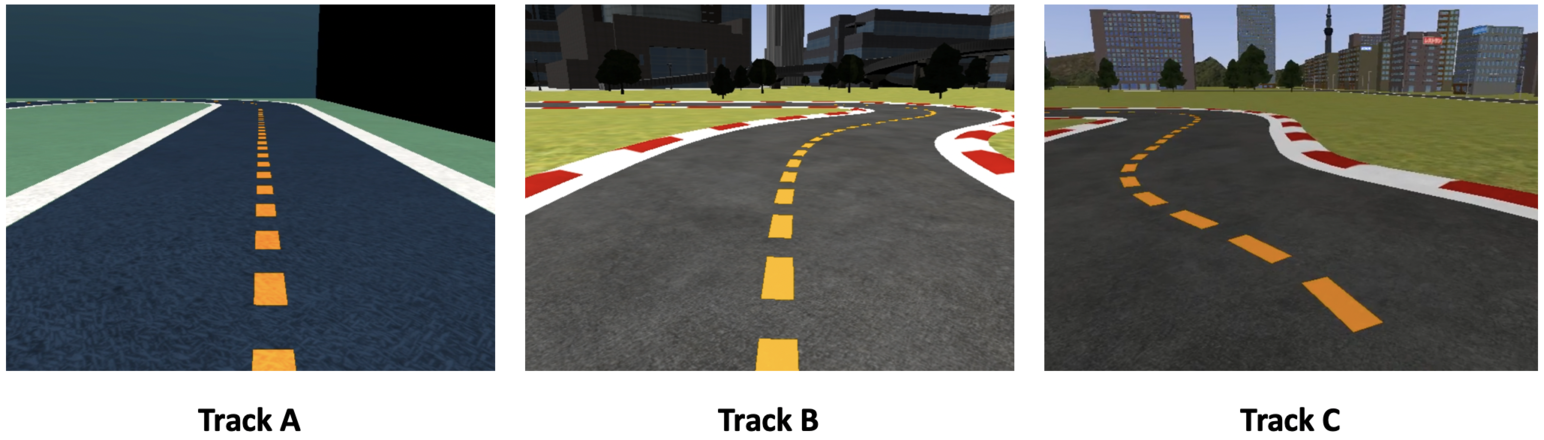}
		\caption{Camera view of simulation tracks}
	\end{subfigure}
	\hfill
	\begin{subfigure}[b]{\columnwidth}
		\centering
		\includegraphics[width=\columnwidth]{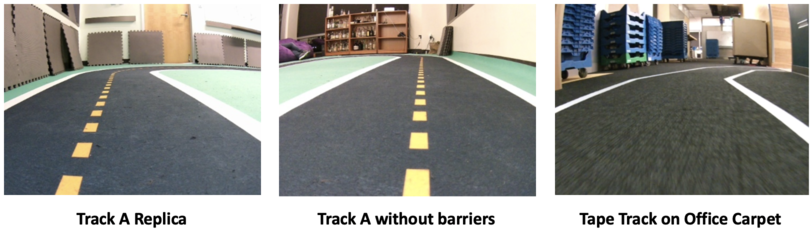}
		\caption{Camera view of real world tracks}
	\end{subfigure}
	\caption{We train in multiple tracks and evaluate with a replica track as well as a track made with duct tape.}
	\vspace{-1em}
\end{figure}


\subsection{Training with Amazon SageMaker}

SageMaker is a platform to train and deploy machine learning models at scale using the Jupyter Notebook~\cite{kluyver2016jupyter} as interface. 
SageMaker integrates RL algorithms using Coach and RLlib~\cite{pmlr-v80-liang18b} libraries that build on top of existing deep learning frameworks. SageMaker uses RL Coach to support the decoupled simulation based training used in DeepRacer, and RLlib for integrated simulation and training. The libraries are packaged in a Docker container~\cite{merkel2014docker} and training can be launched in a cluster of machines with different configurations (CPU/GPU/RAM). The training clusters are created on-demand and billed per second, freeing users from infrastructure maintenance. Metrics such as rewards per episode, the policy entropy, cpu/memory use are visualized, source code is saved and logs are recorded. Users can launch experiments in parallel and search across experiment metadata. In addition to autonomous racing, SageMaker contains RL examples for HVAC control, robot locomotion, portfolio management and more.


\subsection{Simulation with AWS RoboMaker}

RoboMaker is a cloud service to develop, test and deploy robot software. It uses Gazebo for simulation. A \emph{robot model} describes each component of the DeepRacer car - the chassis, wheels, camera, Ackermann steering - their dimensions, how they link together, their properties such as mass and camera angle. We create our tracks and background environment in Blender, a 3D modeling software and import it into Gazebo. We use the ODE physics engine that simulates the laws of physics using the robot model and takes into account factors like collision, friction, acceleration, etc. A rendering engine, OGRE, visualizes the graphics. We use Gazebo plugins to add the camera and light sources. We use ROS~\cite{quigley2009ros} for communication between the agent and the simulation. 
The agent uses ROS to place the car in the track at the beginning of an episode, get images from the camera module, get the car's position, velocity, and send throttle, steering commands to control the car. Users can customize the simulation in Gazebo with their own robot models and environments. 

%
%

\begin{figure}[t]
	\centering
	\includegraphics[width=\linewidth]{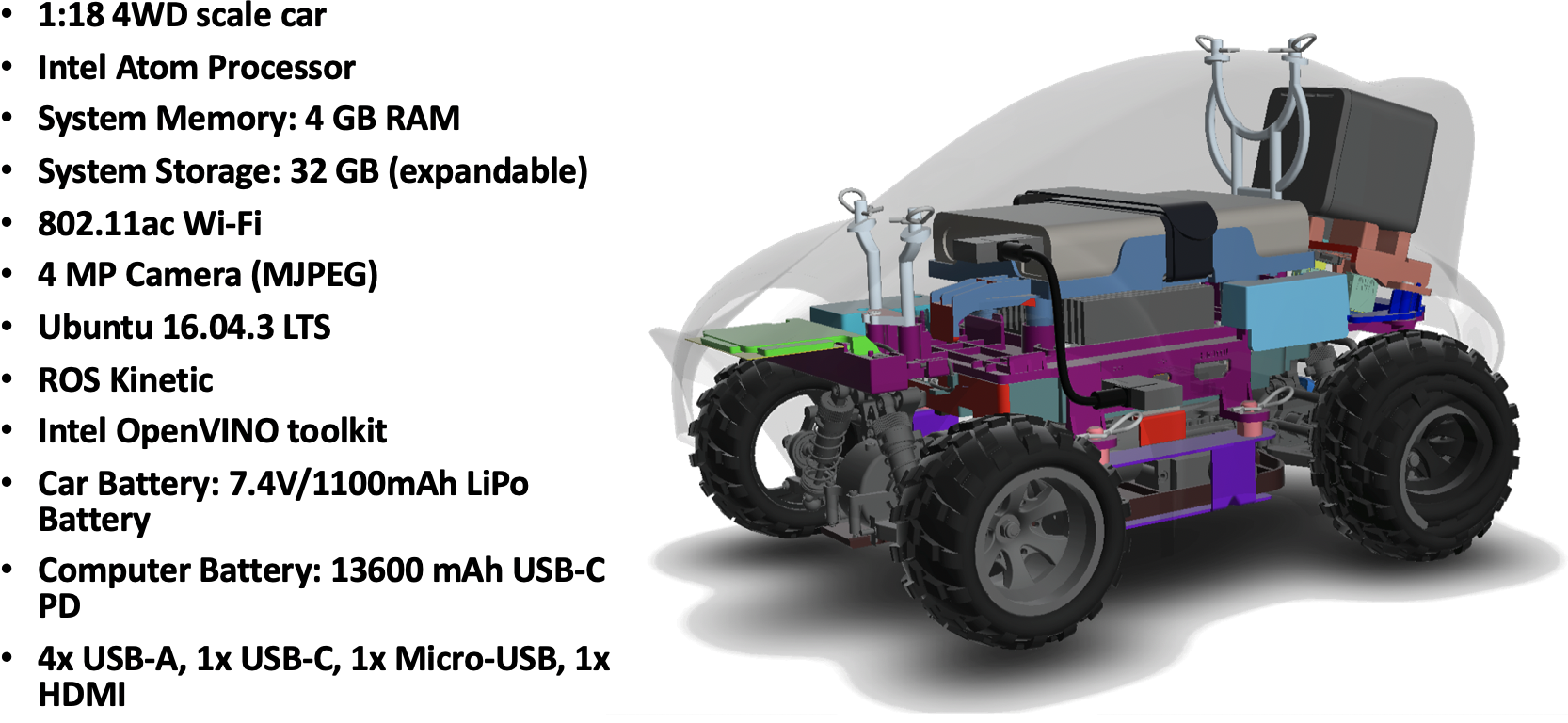}
	\caption{DeepRacer Hardware Specifications}
	\label{fig:hardware}
	\vspace{-1em}
\end{figure}

\subsection{Sim2Real Calibration}


We have matched the URDF robot model to the measured dimensions of the car. We compared images from the real camera and calibrated the height, angle and the field of view of the simulation camera to match the real images. As DeepRacer camera can capture 15 fps, we match the simulation environment to use the same frame rate and use a producer-consumer mechanism to ensure one action per image. We map the agent's action space to the motor control commands by measuring the steering angles and speed of the car under different settings. We have created a real world track that is identical in color, shape and dimensions with one of the simulation tracks. We use barricades around this track to reduce visual distractions. In addition, we have eight other tracks with varying shapes, backgrounds and textures.

\begin{figure*}
	\centering
	\begin{subfigure}[b]{0.32\textwidth}
		\centering
		\includegraphics[width=\textwidth]{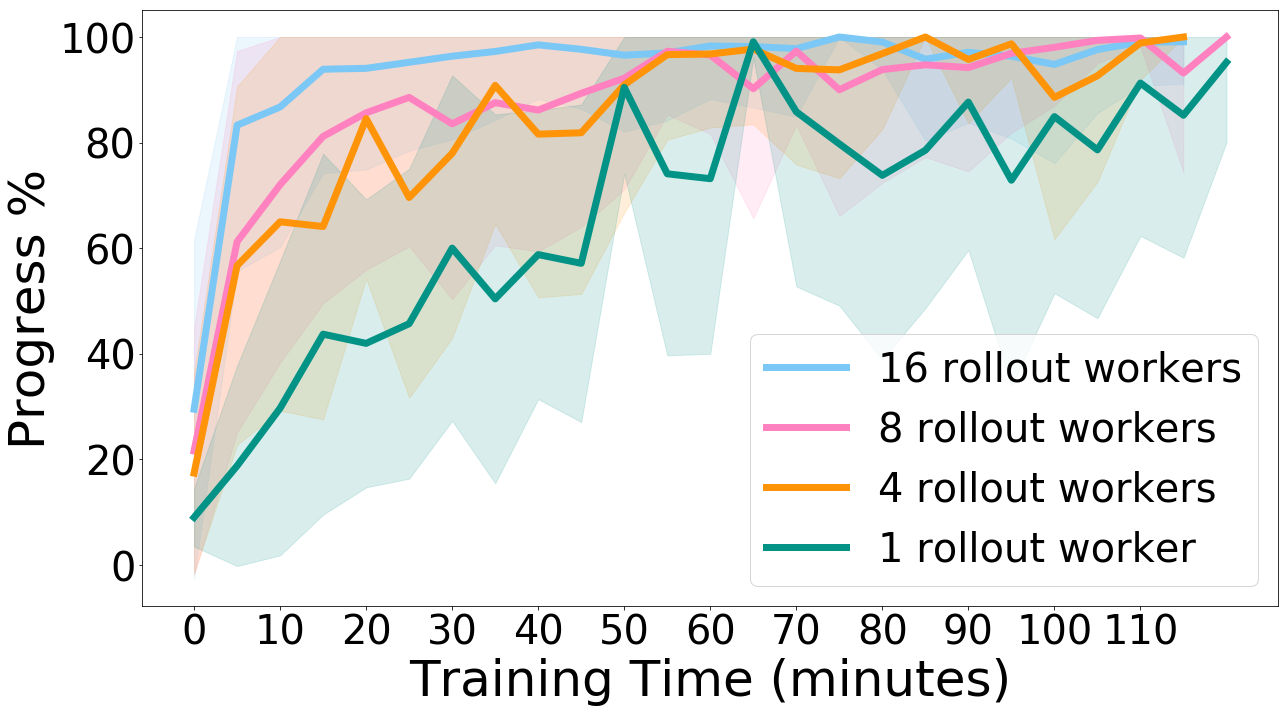}
		\caption{Training with Track A and maximum throttle of 1 m/s}
	\end{subfigure}
	\hfill
	\begin{subfigure}[b]{0.32\textwidth}
		\centering
		\includegraphics[width=\textwidth]{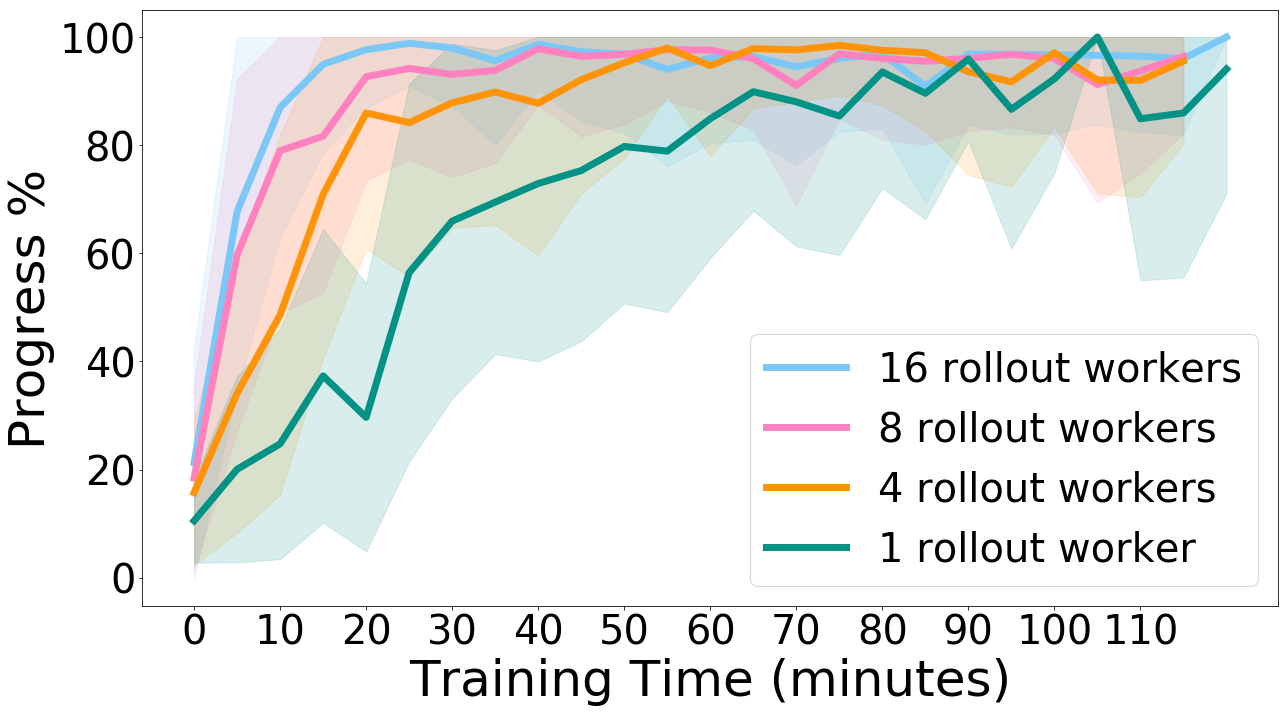}
		\caption{Training with Track A and maximum throttle of 1.67 m/s}
	\end{subfigure}
	\hfill
	\begin{subfigure}[b]{0.32\textwidth}
		\centering
		\includegraphics[width=\textwidth]{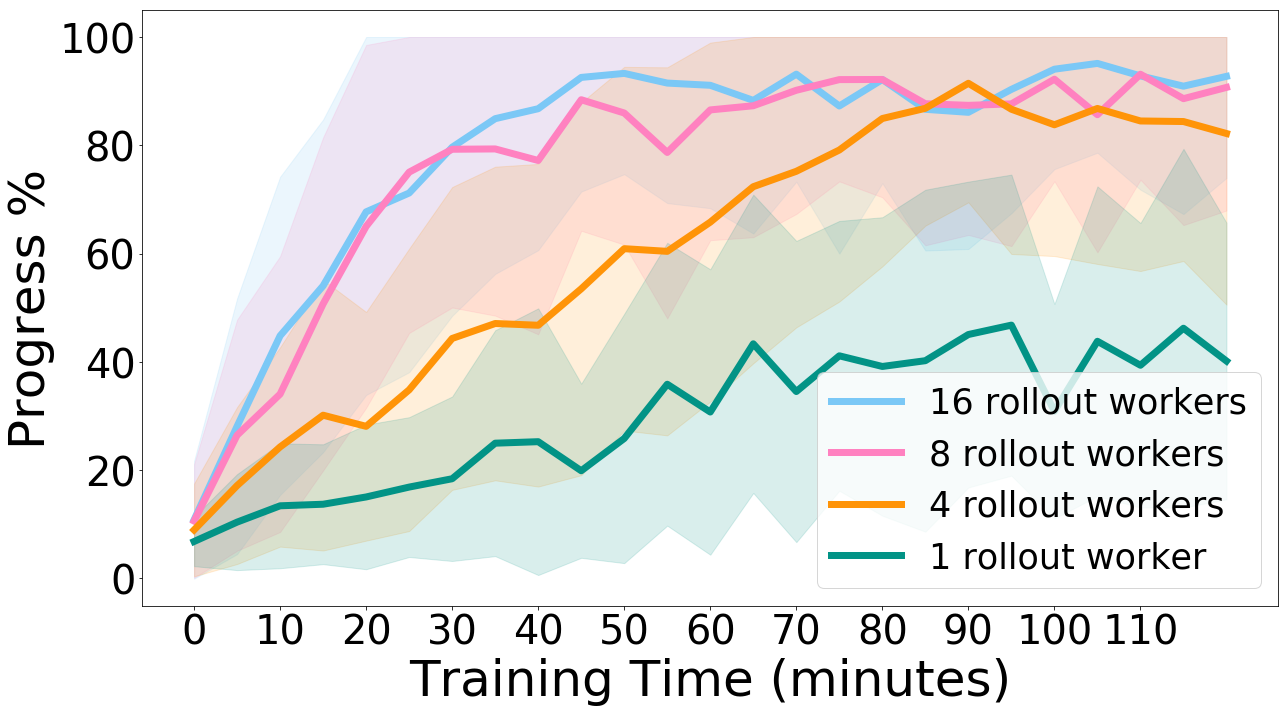}
		\caption{Training with Track B and maximum throttle of 1.67 m/s}
	\end{subfigure}
	\caption{Training with multiple rollout workers. Progress on track is reported across two runs.}
	\label{fig:rollouts}
	\vspace{-1em}
\end{figure*}

\subsection{Calculating Rewards}


We compute an ordered set of points along the middle of the track, called \emph{waypoints}, to estimate the relative position of the car on track. The track and the background are modeled as a polygon mesh. We separate the track mesh from the background and identify the border edges as those which belong to a single triangle. We get two boundaries corresponding to inner and outer part of the track by grouping the border vertices. We construct a bipartite graph from the two sets of vertices and compute the linear sum assignment using the Euclidean distance as edge length. This gives us border vertices parallel to each other on both sizes of the track. The waypoints are the mean of the vertices connected by each edge. The spline is the line joining the waypoints.
The car starts an episode at a waypoint. We flag the car as off-track when it deviates from the spline by more than half the track width. We measure the car's progress by the relative distance it covers compared to the length of the spline.



\subsection{DeepRacer Hardware}

Figure \ref{fig:hardware} gives an overview of DeepRacer hardware. We have designed the car for experimentation while keeping the cost nominal. The Intel Atom processor with a built-in GPU can perform $>$15 inferences per second with our default five layer neural network. The motors are equipped with electronic speed controllers. We can use the car as a regular computer with a monitor, mouse and keyboard connected via HDMI and USB. The camera connects over USB and there are three USB ports for extensions. The 13600 mAh compute battery
 lasts $\sim$6 hours. The 1100 mAh drive battery lasts for $\sim$45 minutes in typical experiments. The WiFi chip enables remote monitoring and programming.
We built the car software on top of ROS. We can load multiple trained models over WiFi. We use Intel OpenVino to convert our Tensorflow models to an optimized binary for fast inference. The camera images are fed to the OpenVino inference engine and a real-time video feed on a browser. There is a web UI for calibrating steering and throttle. The model inference results are converted to motor control commands based on the calibration and action space mapping. In addition, the browser has an interface for manual joystick like control.

\section{Evaluation}
We evaluate our track navigation policies extensively across multiple tracks, with domain randomization in both simulation and real world. We have created a replica of Track A with the track printed on carpet with the same dimensions as in simulation. We place barriers around the track to reduce distractions and evaluate performance both with and without barriers as well as different speeds and lighting conditions. We also made a custom ``tape track'' with 2 inch white duct tape in our office corridor to test model robustness. The track is roughly 24 inches wide, 12m in length, traverses both carpet and concrete, has multiple turns and the car camera is exposed to clutter and bright lights in the background.

\subsection{Training with Multiple Rollouts}

We train policies with three different conditions: on Track A with a maximum throttle of 1 m/s, on Track A with throttle 1.67 m/s and on Track B with throttle 1.67 m/s. The task gets harder at higher speeds. Track B is more difficult to navigate because of background with buildings and higher number of turns. Each episode starts with a different waypoint so that all parts of the track are experienced by the policy. We use p3.2x instance for training in SageMaker and run each experiment twice for 2 hours. Figure \ref{fig:rollouts} shows the progress on track during training with different number of rollout workers.

As we expect, more rollout workers lead to faster convergence. There is diminishing returns as we increase workers, 16 workers give a slightly faster convergence compared to 8. Somewhat surprisingly, the higher throttle of 1.67 m/s helped speedup convergence in Track A. We hypothesize that the agent collects more uniform experience with the faster speed and this helps with convergence. Track B takes longer to converge but follows similar trends as Track A.

\subsection{Robust Evaluation}


\begin{figure*}
	\centering
	\includegraphics[width=\linewidth]{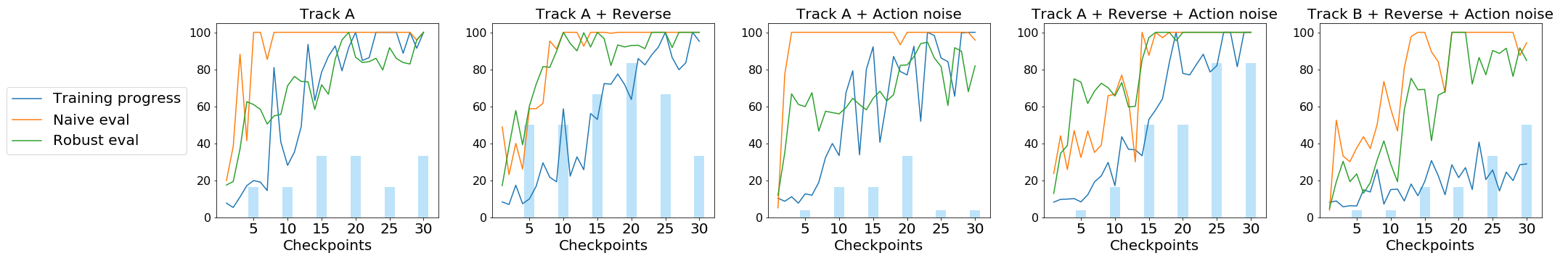}
	\caption{Robust evaluation with domain randomization as a criteria to select policy checkpoints for sim2real transfer. }
	\label{fig:robust_eval}
	\vspace{-1em}
\end{figure*}

\begin{table*}[]
	\caption{Sim2Real for policies trained with regularization and domain randomization. Results are out of 6 trials.}
	\centering
	\label{table:results}
	{%
		\begin{tabular}{|c|c|c|c|c|c|c|c|c|}
			\hline
			\multirow{2}{*}{\makecell{Training\\ Track}} & \multirow{2}{*}{\makecell{Type of \\Training}} & \multirow{2}{*}{\makecell{Checkpoint \# \\(Progress \%)}} & \multicolumn{4}{c|}{Training A Replica} & \multirow{2}{*}{\makecell{Tape Track \\0.7-0.9 m/s}} & \multirow{2}{*}{Total}\\ \cline{4-7}
			& & & 0.5 m/s & 1 m/s & \makecell{Sunlight \\0.8-1 m/s} & \makecell{No Barriers\\ 0.8-1 m/s} & &\\ \hline

			B & \multirow{4}{*}{Default} & 54 (100) & 5 & 3 & 0 & 1 & 3 & 12 \\
			\cline{1-1} \cline{3-9}
			C &  & 53 (99.7) & 5 & 3 & 2 & 3 & 3 & 16 \\
			\cline{1-1} \cline{3-9}
			D &  & 50 (100) & 5 & 3 & 3 & 3 & 0 & 13 \\
			\cline{1-1} \cline{3-9}
			\cline{2-9}
			& L2=2e-5 & 53 (100) & 5 & 4 & 2 & 4 & 2 & 17 \\
			\cline{2-9}
			& Dropout=0.3 & 49 (100) & \textbf{6} & 3 & 5 & 5 & 4 & 23 \\
			\cline{2-9}
			& BatchNorm & 41 (100) & 4 & 2 & 1 & 4 & 2 & 13 \\
			\cline{2-9}
			& Throttle=0.33 m/s & 21 (100) & 2 & 0 & 0 & 0 & 2 & 4 \\
			\cline{2-9}
			& Throttle=1.67 m/s & 72 (91.1) & \textbf{6} & 4 & 5 & \textbf{6} & 2 & 23 \\
			\cline{2-9}
			& Throttle=2.33 m/s & 79 (57.9) & \textbf{6} & 5 & 5 & \textbf{6} & 2 & 24 \\ \hline
			\multirow{7}{*}{B, D} & Default & 41 (100) & 3 & 3 & 3 & 3 & 1 & 13 \\
			\cline{2-9}
			& Color Aug. & 49 (100) & \textbf{6} & 5 & \textbf{6} & \textbf{6} & 3 & 26 \\
			\cline{2-9}
			& Translation & 37 (100) & \textbf{6} & 5 & 5 & 3 & 3 & 22 \\
			\cline{2-9}
			& Shadow & 46 (100) & 5 & 3 & 5 & 3 & 2 & 18 \\ \cline{2-9}
			& Sharpen & 48 (89.5) & 4 & 4 & 5 & 4 & 0 & 17 \\ \cline{2-9}
			& Pepper & 53 (98.9) & \textbf{6} & 3 & 4 & 2 & 1 & 16 \\ \cline{2-9}
			& All image aug & 48 (100) & 5 & \textbf{6} & 3 & 4 & 0 & 18 \\ \hline
			C & \begin{tabular}[c]{@{}c@{}}Best combo,\\ Throttle=2.33 m/s\end{tabular} & 67 (91.7) & \textbf{6} & \textbf{6} & \textbf{6} & 5 & 4 & \textbf{27} \\ \hline
		\end{tabular}%
	}
	\vspace{-1em}
\end{table*}

We test whether robust evaluation in simulation is indicative of real world performance. If true, we can identify when to stop training in simulation and avoid underfitting/overfitting. We can tune our hyper-parameters entirely in simulation and avoid extensive testing in the real world. We train policies with increasing levels of domain randomization and evaluate the policy in both simulation and real.

Our baseline case is trained on Track A with no domain randomization and throttle of 1 m/s. For domain randomization, we train policies on Track A with (i) up to 10\% uniform random noise to steering and throttle (action noise), (ii) reverse direction of travel each episode (reverse), (iii) include both action noise and reverse, and (iv) train on Track B with both action noise and reverse. For robust evaluation, we add uniform random noise to actions, evaluate in multiple starting positions and both directions of travel on Track A. For naive evaluation, we evaluate on Track A with a fixed starting point without randomization. Both evaluations test each checkpoint 10 times in simulator. We pick six policies during training from checkpoints 5 through 30, and test their sim2real performance in the Track A replica with 3 trials for each direction of travel. The model performance varies with speed, but it is difficult to maintain a constant speed due to changing battery levels and as the model switches between throttle levels. For sim2real experiments we ensure the model completes a lap in 18 to 22 seconds (0.8-1 m/s). In simulation, the models complete the lap in $\sim$35 seconds, so we test the policy at about double speed in the real track.

Figure \ref{fig:robust_eval} shows the experimental results. The model that perform consistently well with robust evaluation also perform well on the real track. The models are particularly robust when a sequence of checkpoints perform well in simulator. Reversing the direction of travel significantly improves model performance. Action noise does not help by itself, but improves performance when combined with reverse. Policies trained on Track B do not perform well for checkpoints in Figure \ref{fig:robust_eval}, but with more training start performing well in both robust evaluation and real track, policy checkpoint 35 traversed the real track successfully 5 out of 6 trials.

The performance of the model changes dramatically at slower speeds (35s lap, 0.5 m/s), even checkpoint 5 of the policy trained on Track A with no randomization traverses the real track. This model is trained in \textless5 minutes. All the above policies were trained in \textless1 hour with 4 rollouts.

\subsection{Robust Sim2Real}

We test the robustness of sim2real by training on multiple tracks, with multiple speeds, regularization and domain randomization in actions and observations. By default, we train on Track B with throttle of 1 m/s, with action noise and reverse direction each episode. We pick model checkpoints based on performance in robust evaluation and test the policy on Track A replica in two speeds (0.5 m/s, 1 m/s), with bright sunlight, with no barriers and on tape track.

Table \ref{table:results} summarizes our results. Training on a different track gives good sim2real results, but vary track to track. 
For regularization, we used L2 norm, dropout, batch normalization and an entropy bonus to the policy loss.  We tested the models that give best performance in robust evaluation. Reducing the entropy bonus to 0.001 (it is 0.1 by default) and dropout with probability 0.3 were particularly effective. Larger throttle speeds in training increased the robustness of the model dramatically but also increased convergence time in the presence of action noise. Mixing multiple tracks during training did not lead to improvement in performance. We perturb the observation images with random color, horizontal translation, shadow, and salt and pepper noise, each with 0.2 probability. For random color, we combine the effects of random hue, saturation, brightness and contrast to create variations in observation. Random color was the most effective method for sim2real transfer.

We combine the best of our parameters and train a model on Track C with L2 regularlization, lower entropy bonus, dropout, color randomization and a maximum throttle of 2.33 m/s. This model performed the best overall in our experiments. The model consistently completed 11 second laps (1.6 m/s) in our Track A replica.

\section{Conclusion}
DeepRacer is an experimetation and educational platform for sim2real reinforcement learning. The platform integrates state-of-the-art Deep RL algorithms, multiple simulation engines with OpenAI Gym interface, provides on-demand compute, distributed rollouts that facilitates domain randomization and robust evaluation in parallel. We demonstrate DeepRacer platform features with a 1/18th scale car that navigates a race track using reinforcement learning. We have created a calibrated robot model for the car in Gazebo along with multiple race tracks. We demonstrate robust sim2real navigation performance trained in DeepRacer with PPO algorithm in both our real world replica track as well as a custom tape track. We achieve sim2real in real track with \textless5 minutes of training at slow speeds and achieve speeds of 1.6 m/s using models trained with tuned parameters. Thousands of users have replicated our model training and demonstrated sim2real RL navigation.

\bibliographystyle{IEEEtran}
\bibliography{refs}

\end{document}